\documentclass[10pt,twocolumn,letterpaper]{article}

\usepackage{cvpr}
\usepackage{graphicx}
\usepackage{amsmath}
\usepackage{amssymb}
\usepackage{booktabs}
\usepackage{multirow}
\usepackage{array}
\usepackage{xcolor}

\usepackage{xspace}
\usepackage[pagebackref,breaklinks,colorlinks]{hyperref}
\usepackage[capitalize]{cleveref}
\crefname{section}{Section}{Sections}
\Crefname{section}{Section}{Sections}
\crefname{table}{Table}{Tables}
\Crefname{table}{Table}{Tables}
\crefname{figure}{Figure}{Figures}
\Crefname{figure}{Figure}{Figures}

\def\cvprPaperID{*****}
\def\confName{CVPR}
\def\confYear{2026}

\newcommand{\PlaySyn}{PlaySyn\xspace}
\newcommand{\PlayBench}{PlayBench\xspace}
\newcommand{\MossPlay}{MOSS-Image-Edit-PlaySyn\xspace}
\newcommand{\QwenPlay}{Qwen-Image-2.1-PlaySyn-LoRA\xspace}
\newcommand{\field}[1]{\textit{#1}}

\definecolor{CreativeHook}{RGB}{0,128,113}
\definecolor{CheckableDetail}{RGB}{53,94,158}
\newcommand{\CBhook}[1]{\textcolor{CreativeHook}{\textbf{#1}}}
\newcommand{\CBdetail}[1]{\textcolor{CheckableDetail}{#1}}
\makeatletter
\renewcommand\paragraph{\@startsection{paragraph}{4}{\z@}
  {3.25ex \@plus1ex \@minus.2ex}{-1em}{\elvbf}}
\makeatother

\begin{document}

\title{PlaySyn: Mining Creative Patterns for Image Editing Data and Evaluation}

\author{
Yuxiang Jiang\textsuperscript{*}\quad
Cong Guo\textsuperscript{*}\quad
Ke Lin\textsuperscript{\dag}\quad
Youzheng Wu\quad
Xiaodong He\\
JD Technology\\
{\tt\small \{jiangyuxiang2,guocong9,linke.jason,wuyouzheng,hexiaodong\}@jd.com}
}

\twocolumn[{%
  \maketitle
  \centering
  \includegraphics[width=\textwidth]{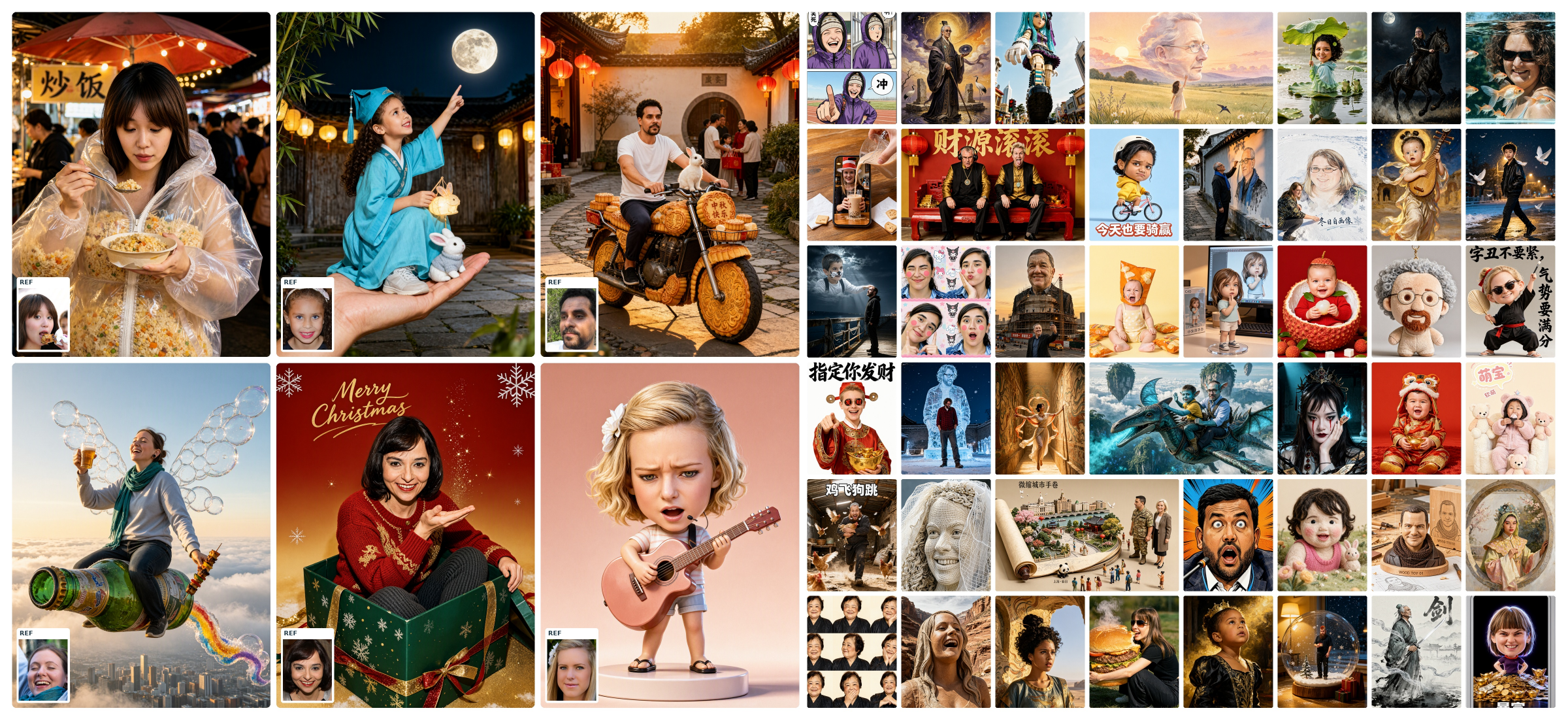}
  \captionof{figure}{Overview of \PlaySyn-100K. The dataset contains 105,724 quality-controlled and human-curated training samples for creative image editing. Large images are synthesized targets and small insets are reference images. The examples cover reusable creative tasks, single and multi-reference inputs, identity preservation, and diverse visual expression.}
  \label{fig:gallery}
  \vspace{1em}
}]
\begingroup
\renewcommand{\thefootnote}{\fnsymbol{footnote}}
\footnotetext[1]{Equal contribution.\quad\textsuperscript{\dag}Project Lead.}
\endgroup

\begin{abstract}
Creative image editing has two failure modes: an output can satisfy every instruction and still look dull, or look striking and lose the person’s identity. Existing editing datasets mostly cover atomic operations, so they rarely capture recurring creative patterns. We build \PlaySyn-100K, 105,724 reference-guided samples from patterns mined from about 1.5M real edits. For each pattern, we record a CreativeDNA---visual hook, contrast, story, required edits, and reference roles---and turn it into a CreativeBrief that fixes what must change and what must be preserved. We instantiate each brief with new references, generate candidates, and keep only those that pass fidelity, identity, anatomy, OCR, and quality checks. \PlayBench adds 2,050 frozen tasks with Pointwise hard gates and position-randomized Pairwise comparison. On \PlayBench, \MossPlay and \QwenPlay are both preferred over Qwen-Image-2.1 for image quality and interestingness. Mining real edits for reusable patterns, rather than using the raw interactions as training pairs, is therefore a useful source of supervision for creative editing.
\end{abstract}

\section{Introduction}
\label{sec:intro}

Creative image editing can turn an ordinary photo into a movie poster, place a person alongside a favorite character, or construct new stories from familiar people and settings (\Cref{fig:gallery}). In these examples, the editing operation alone does not specify the desired visual concept.

We use \emph{play pattern} to denote a recurring creative editing intent, including its visual mechanism, reference roles, and preservation constraints. Such patterns are difficult to recover from real interactions because user intents are expressed at different granularities and recur in many related forms. Existing datasets and benchmarks largely target semantic edits, object replacement, and local control~\cite{zhang2023magicbrush,sheynin2024emu}. A canonical operation such as background replacement does not specify the visual contrast or story that distinguishes one creative task from another, and a longer prompt cannot recover information missing from the template. Evaluation must also account for both constraint satisfaction and interestingness: an attractive image may lose the person's identity, while an image that completes every edit may still look ordinary.

We mine CreativeDNA from real creative patterns and convert it into CreativeBriefs for new references. Each task specifies reference roles, identity sources, changes, and preserved attributes. Candidate synthesis is followed by fidelity checking, two-stage quality control, and interestingness assessment. \PlayBench holds out frozen tasks from training and divides them into Core, Challenge, and Creative tracks. Its Pointwise evaluation records constraint satisfaction, while Pairwise evaluation compares two outputs on the same task.

Our main contributions are:
\begin{itemize}
  \item We introduce \PlaySyn, a dataset and mining pipeline built around recurring creative editing patterns. Starting from real editing interactions, the pipeline discovers patterns, instantiates them with new references, synthesizes candidates, and applies hard quality filtering, yielding 105,724 training samples.
  \item We construct \PlayBench, a frozen benchmark with 2,050 tasks across 35 domains, combining Pointwise quality gates with randomized Pairwise evaluation.
  \item We adapt the text-to-image MOSS-Image model to multi-reference editing through dual frozen encoding paths and in-place VAE-token replacement. Experiments with MOSS PlaySyn and Qwen PlaySyn LoRA show consistent Pairwise quality and interestingness gains over Qwen-Image-2.1.
\end{itemize}

\begin{figure*}[t]
  \centering
  \includegraphics[width=\textwidth]{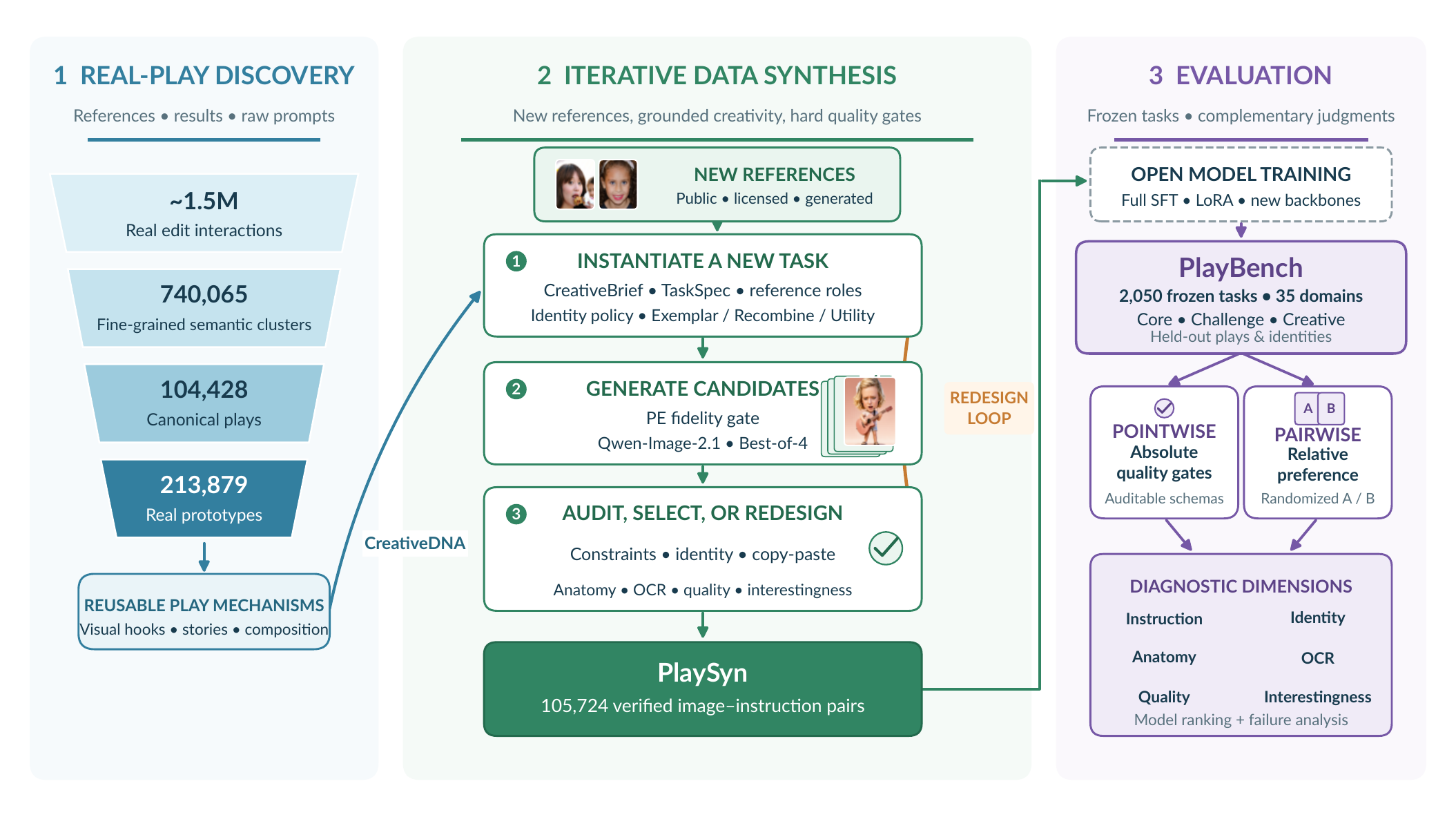}
  \caption{\PlaySyn pipeline and \PlayBench. (1) We mine $\sim$1.5M real edits into 740K semantic clusters, 104K canonical patterns, and 214K CreativeDNA records. (2) Each CreativeDNA plus new references becomes a task; candidates are generated, checked for fidelity, filtered by two hard-quality stages, and ranked for interestingness. Failed cases are diagnosed and resynthesized. (3) \PlayBench is frozen: Pointwise hard gates plus randomized Pairwise preferences.}
  \label{fig:overview}
\end{figure*}

\section{Related Work}
\label{sec:related}

\subsection{Instruction-Based Image-Editing Data}

Instruction-based image-editing data come from human annotation, rule-based transformations, and generative synthesis. MagicBrush provides human-annotated edits~\cite{zhang2023magicbrush}, whereas InstructPix2Pix synthesizes training pairs with a language model and a text-to-image model~\cite{brooks2023instructpix2pix}. Emu Edit combines recognition and generation tasks for precise editing~\cite{sheynin2024emu}. HQ-Edit and OmniEdit use multimodal models to construct or filter editing data~\cite{hui2024hqedit,wei2024omniedit}, and SEED-Data-Edit and GPT-Image-Edit-1.5M explore hybrid and large-scale GPT-generated corpora~\cite{ge2024seeddataedit,wang2025gptimageedit}.

Scalable pipelines often pair task expansion with automatic verification. UltraEdit uses real-image anchors and LLM-based expansion~\cite{zhao2024ultraedit}, and AnyEdit covers a broad set of local and global operations~\cite{yu2025anyedit}. Step1X-Edit and ImgEdit add real-user benchmarks, high-resolution references, and multi-turn tasks~\cite{liu2025step1xedit,ye2025imgedit}. NoHumansRequired, ScaleEdit-12M, and UnicEdit-10M study autonomous mining, multi-agent generation, and unified verification~\cite{kuprashevich2025nhr,chen2026scaleedit,ye2025unicedit}. Pico-Banana-400K includes multi-turn and preference subsets~\cite{qian2026picobanana}, while ConceptEdit-12M organizes data around fine-grained concepts and compositional edits~\cite{cui2026conceptedit}. \PlaySyn organizes data around the visual mechanisms and stories that recur in real creative edits.

\subsection{Evaluating Creative Image Editing}
\label{sec:creative_eval}

Image-editing evaluation typically measures instruction following, content preservation, semantic consistency, and visual quality. GEdit-Bench grounds general editing evaluation in real user instructions~\cite{liu2025step1xedit}; UnicBench additionally examines non-edit consistency and spatial or knowledge-driven reasoning~\cite{ye2025unicedit}; and ConceptEdit-Bench diagnoses capabilities across fine-grained editing concepts~\cite{cui2026conceptedit}. As editing moves from deterministic object modifications to open-ended creative generation, a single correctness metric cannot fully characterize output quality. We predefine Utility and Creative task types and interpret their value differently: Utility tasks are not penalized for lacking novelty, whereas Creative tasks emphasize novelty, composition, storytelling, and shareability.

\subsection{Multimodal Models as Judges}

Multimodal large models can jointly reason about reference images, natural-language instructions, and generated outputs, making them scalable judges for complex image editing. Their judgments can be affected by position, rating scale, and prompt design~\cite{zheng2023judging}. We use structured atomic judgments in Pointwise evaluation, randomize left--right positions in Pairwise evaluation, and distinguish automatic judgments from human calibration.

\section{PlaySyn Data Construction}
\label{sec:data}

\subsection{Collecting and Normalizing Real Play Patterns}

We start from approximately 1.5M real image-editing interactions. Raw prompts and edit descriptions inferred from image pairs are used to extract pattern names, editing templates, operations, and output formats. The full taxonomy is induced from all samples. Coarse partition-level clustering, semantic clustering within partitions, and subdivision of large clusters produce 740,065 fine-grained clusters. We normalize names separately for clusters with at least 10 instances and those with 3--9 instances, then deduplicate across partitions to obtain 104,428 real creative patterns. They cover approximately 55\% of real instances; singletons and doubletons remain in a long-tail pool outside the normalized library.

\begin{figure*}[t]
  \centering
  \includegraphics[width=\textwidth]{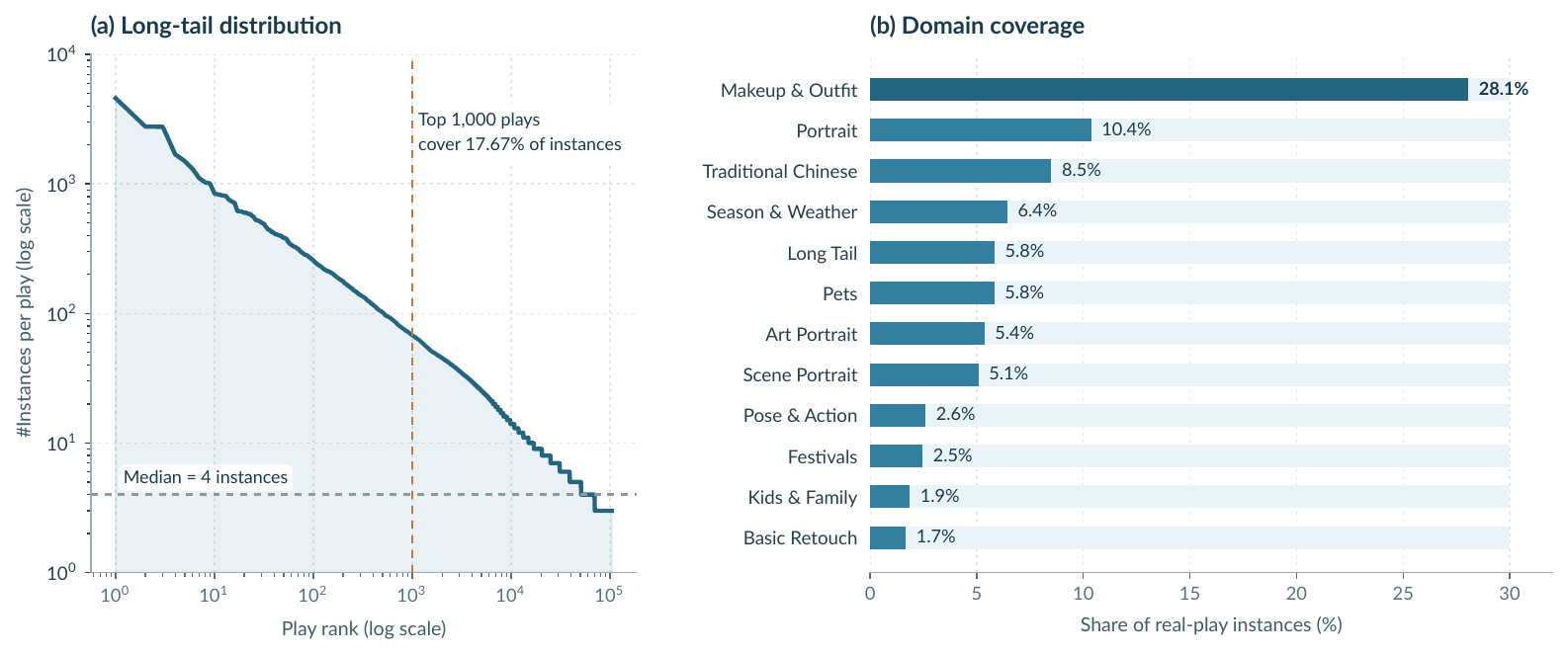}
  \caption{Real-play distribution and domain coverage. (a) Long-tail distribution: the median is four instances per pattern; the top 100 and top 1,000 plays cover 6.25\% and 17.67\% of instances. (b) Share of real-play instances in each of the 12 domains.}
  \label{fig:real_play_analysis}
\end{figure*}

Real creative patterns exhibit a strong long tail. The median number of instances per pattern is four, and the 90th and 99th percentiles are 14 and 66. The top 100 and top 1,000 creative patterns account for only 6.25\% and 17.67\% of cataloged instances, respectively.

Makeup and outfit changes form the largest domain, accounting for 28.1\% of instances, followed by portraits (10.4\%) and traditional Chinese styles (8.5\%). The other domains range from seasonal scenes and artistic portraits to pets, festivals, and family activities, covering both popular and less frequent forms of creative editing.

Real interactions are used only for pattern mining; training and evaluation use newly instantiated samples (\Cref{tab:data_usage_boundary}).

\subsection{Creative Mining and Structured Representation}
\label{sec:creative_mining}

\paragraph{Canonical templates.} Canonical templates support play clustering, data isolation, reference assignment, and quality checking, but they often reduce diverse samples to generic operations such as identity preservation, outfit changes, background replacement, and pose adjustment. This compression discards visual contrast and narrative context, which longer prompts or adjusted Prompt Enhancement (PE) parameters cannot recover. ConceptEdit studies distributional collapse caused by random sampling from coarse categories~\cite{cui2026conceptedit}. Our focus is information loss within each operation: one canonical operation may cover many distinct visual hooks and stories.

\paragraph{Task representation.} Each task contains a TaskSpec and a CreativeBrief. TaskSpec $T=(\tau,\mathcal{C},\mathcal{P},\mathcal{R},\mathcal{I})$ contains an editing template $\tau$, atomic change and output constraints $\mathcal{C}$, preserved attributes $\mathcal{P}$, reference roles $\mathcal{R}$, and identity sources $\mathcal{I}$. CreativeBrief describes visual hooks, story, contrast, composition, and emotional value. During task construction, new creative requirements are added to TaskSpec as verifiable constraints. \Cref{sec:supp_schema,sec:supp_examples} provide the structured schemas and conversion examples.

\paragraph{CreativeDNA.} For each pattern, we sample prototypes round-robin from its pre-clustering subcategories, so one prototype does not stand for only one narrow subcategory. For the 71,293 creative patterns entering the synthesis pipeline, we retrieve up to three real prototypes each, obtaining 213,879 prototypes, of which 97.72\% originate from real user data. Prototype $j$ is written as $E_j=(\mathcal{R}_j,I_j^{\mathrm{out}},p_j,c_j)$, where $\mathcal{R}_j$ is the set of real reference images, $I_j^{\mathrm{out}}$ is the output image from the original editing interaction, $p_j$ is the raw prompt, and $c_j$ is a detailed edit description. Given $(\mathcal{R}_j,I_j^{\mathrm{out}},p_j,c_j)$, a VLM writes one CreativeDNA record $D_j$. We keep multiple $D_j$ per pattern when real samples realize it differently.

Each DNA record stores four kinds of information:
\begin{itemize}
  \item \textbf{Task type and interestingness:} \field{play\_type} distinguishes Creative from Utility; \field{interest\_score} rates interestingness on a 1--5 scale rather than image quality. Utility records must provide \field{utility\_note}, while Creative records require a nonempty visual hook.
  \item \textbf{Visual hook, contrast, and story:} \field{visual\_hook} records a concrete visible point of interest; \field{contrast} describes contrasts in scale, era, material, or reality--fantasy; \field{story} and \field{subject\_relation} describe the depicted event and relationships among subjects.
  \item \textbf{Visual elements, composition, and emotion:} \field{signature\_elements} lists key visual elements, \field{composition} records layout and viewpoint, and \field{emotional\_payoff} describes surprise, humor, comfort, awe, or other emotional value.
  \item \textbf{Required edits and references:} \field{required\_operations} lists edits required to realize the mechanism, and \field{reference\_requirements} records the prototype's input requirements. Missing fields, empty arrays, invalid types, and out-of-range scores fail structural validation.
\end{itemize}
We do not treat quality words—high-definition, premium, beautiful, atmospheric—as creativity. A Creative record must name a visible hook and an event; a Utility record such as an ID photo or retouch keeps its practical goal and is not forced to be novel. In this prototype collection, 90.81\% are classified as Creative and 9.19\% as Utility.

\begin{table*}[t]
  \centering
  \footnotesize
  \setlength{\tabcolsep}{3.5pt}
  \hyphenpenalty=10000
  \caption{Comparison of representative instruction-based image-editing datasets.}
  \begin{tabular}{>{\raggedright\arraybackslash}p{0.125\textwidth}l>{\raggedright\arraybackslash}p{0.14\textwidth}>{\raggedright\arraybackslash}p{0.15\textwidth}l>{\raggedright\arraybackslash}p{0.10\textwidth}}
    \toprule
    Dataset & Scale & Reference source & Task source & Focus & Resolution \\
    \midrule
    MagicBrush~\cite{zhang2023magicbrush} & 10,388 & MS COCO & Human annotation & Local & 500 px \\
    UltraEdit~\cite{zhao2024ultraedit} & $\sim$4M & Real images & LLM expansion & Local/general & 512 px \\
    AnyEdit~\cite{yu2025anyedit} & 2.5M & Real/synthetic images & Counterfactual edits & General & 512 px \\
    ImgEdit~\cite{ye2025imgedit} & 1.2M & LAION-Aesthetics; videos & Auto-generated tasks & Local/general & Short side $\geq1280$ px \\
    Pico-Banana-400K~\cite{qian2026picobanana} & $\sim$400K & OpenImages & 35-type taxonomy & Portrait & -- \\
    ConceptEdit-12M~\cite{cui2026conceptedit} & 12M & Synthetic images & Concept expansion & Local/general & Mostly 2 MP \\
    \midrule
    \PlaySyn\ (ours) & 105,724 & FFHQ; expanded portraits & Creative-pattern mining & Creative generation & 4 MP \\
    \bottomrule
  \end{tabular}
  \label{tab:dataset_comparison}
\end{table*}

\paragraph{CreativeBrief.} For each new instance, the model reads the user instruction, canonical template, base constraints, preserved attributes, reference roles, and descriptions of the new references. It also reads all valid DNA records from the same play. Using these inputs, it generates two Briefs in one pass. Each Brief records the creative mechanism, additional constraints, and a natural user instruction called \field{creative\_prompt}. The two Briefs must differ in mechanism, story, or composition. For a Creative task, the \textbf{Exemplar} Brief inherits the core mechanism of a single prototype. The \textbf{Within-play Recombine} Brief mixes compatible mechanisms from the same play. For a Utility task, the model produces two proposals that improve practical value. The reference count and roles remain fixed, and every \texttt{<Picture N>} must appear in \field{creative\_prompt}.

\paragraph{Constraints and identity.} \field{additional\_constraints} contains only atomic, visible requirements beyond the base constraints. It does not repeat identity-preservation requirements or generic quality terms, and each Brief adds at most 10 constraints. The $k$-th new requirement is stored as \field{creative\_c}$_k$, with \field{type=change}, \field{required=true}, and its associated reference indices. The final task satisfies
\begin{equation}
  \begin{aligned}
    \mathcal{C}_{\mathrm{task}}&=\mathcal{C}_{\mathrm{base}}\cup\mathcal{C}_{\mathrm{additional}},\\
    \mathcal{P}_{\mathrm{task}}&=\mathcal{P}_{\mathrm{base}},\\
    \mathcal{R}_{\mathrm{task}}&=\mathcal{R}_{\mathrm{base}}.
  \end{aligned}
\end{equation}
\field{creative\_prompt} becomes the new user instruction, while the original remains \field{base\_prompt} for traceability. The identity policy then rechecks the task and template, excluding tasks that contain ambiguous identity references or failed policies. The full pipeline produces 142,570 valid Briefs: 67,775 Exemplar, 67,775 Within-play Recombine, and 7,020 Utility-grounded.

\paragraph{Conversion example.} For an exaggerated big-head portrait sitting on a money bag, the real DNA breaks into three parts: a contrast between realistic skin texture and cartoonish head proportions; a subject relationship where the person is partly surrounded by the money bag and banknotes; and a two-part composition with the head above and the bag below. Required operations include enlarging the head, changing to a seated laughing pose, and adding a burlap money bag and banknotes while preserving identity and clothing. For a new portrait, the Exemplar Brief keeps this mechanism and adds a few small, specific constraints: banknotes fan out from both sides of the bag opening, gold coins are scattered nearby, the burlap weave is visible, and the enlarged head has realistic skin. A Within-play Recombine treats the money bag as a cradle and adds piles of banknotes and coins and a red-and-gold festive set. Quality control can then verify each creative element in the output.

\subsection{Reference Instantiation and Identity Policy}

Our portrait references come from FFHQ~\cite{karras2019stylegan}. We remove non-single-person and low-quality images, annotate age group, gender, expression, head pose, and clothing, and split the pool into close-up, half-body, and full-body images with default proportions of approximately 50\%, 30\%, and 20\%. Close-ups retain the FFHQ originals. Qwen-Image-2.1~\cite{qwen2026image21} extends the other images to half- or full-body shots. It preserves the face, body proportions, skin tone, and hair, while adding clothes, pose, and background. Each task assigns its references explicit roles, such as target canvas, identity source, style source, or layout source, and records whether identity must be preserved. Identity replacement and multi-person interaction tasks get additional checks for role confusion, unsupported identity blending, and ambiguous references. Multireference tasks must combine the input images according to each image’s role.

\begin{figure*}[t]
  \centering
  \includegraphics[width=\textwidth]{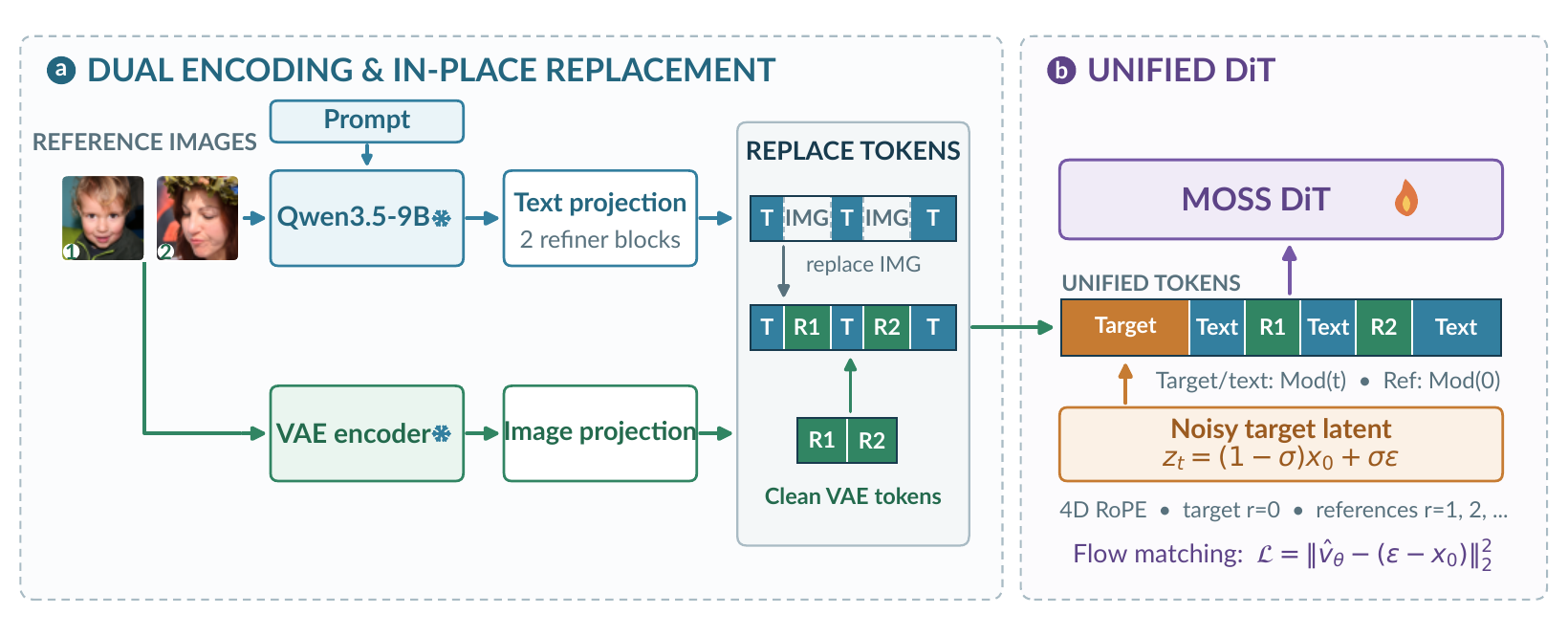}
  \caption{Architecture of MOSS-Image-Edit-PlaySyn. Reference images follow two frozen encoding paths. After text projection, VAE tokens replace VLM image placeholders in place. The replaced context is concatenated with noisy target tokens and processed by joint attention in the MOSS DiT. Clean references use $\mathrm{Mod}(0)$, while target and text tokens use $\mathrm{Mod}(t)$. The VAE and Qwen encoder are frozen; the full DiT is trained.}
  \label{fig:moss_playsyn_architecture}
\end{figure*}

\subsection{Prompt Enhancement and Candidate Synthesis}

Training keeps a natural, concise \field{prompt}, while candidate synthesis uses a \field{generation\_prompt} with more visual detail. The former is the user-facing instruction; the latter is used only for candidate generation. A PE rewrite cannot directly trigger image generation. The system checks it against the original instruction, all references, TaskSpec, and CreativeBrief, including visual hooks, changes, preserved attributes, identity roles, specified text, counts, and spatial relationships. It also distinguishes details already present in a reference from unsupported additions. Failed rewrites receive only the necessary minimal repairs and are checked again with images; only the current passing version may enter generation, as detailed in \cref{sec:supp_pe}. Each original task has two Briefs, each generated with two seeds to produce four candidates. After synthesis by Qwen-Image-2.1~\cite{qwen2026image21}, candidates are selected by constraints, identity, quality, and structural correctness.

\subsection{Two-Stage Quality Control and Interestingness-Based Curation}

The first quality-control stage checks atomic changes, preserved attributes, identity consistency, direct copying or crude compositing, and severe quality problems. A second independent review examines head--body proportions, finger count and ownership, extra limbs, cross-person limb merging, facial corruption, instruction completion, and text. Interestingness ranking uses the dimensions in \Cref{sec:creative_eval}, with labels for plain edits, Utility play, and Creative play. It is applied only after the hard gates and cannot override identity or core-constraint failures. Samples requiring deletion or redesign are excluded from the final training data. The resulting \PlaySyn training dataset contains 105,724 samples: 77,371 \field{keep\_creative}, 15,530 \field{keep\_utility}, and 12,823 \field{review}.

\subsection{Comparison with Existing Image-Editing Datasets}

\Cref{tab:dataset_comparison} compares reference sources, task organization, and resolution. Most existing datasets scale editing data by expanding operations or filtering synthetic candidates. UltraEdit and AnyEdit broaden operation coverage, ImgEdit emphasizes resolution, and ConceptEdit-12M expands fine-grained concepts. These datasets are organized by operation or concept; \PlaySyn is organized by recurring creative mechanisms mined from real edits. Each sample links visual and narrative mechanisms to reference roles and verifiable constraints, with hard quality evaluation separated from interestingness assessment.

\begin{figure*}[t]
  \centering
  \includegraphics[width=\textwidth]{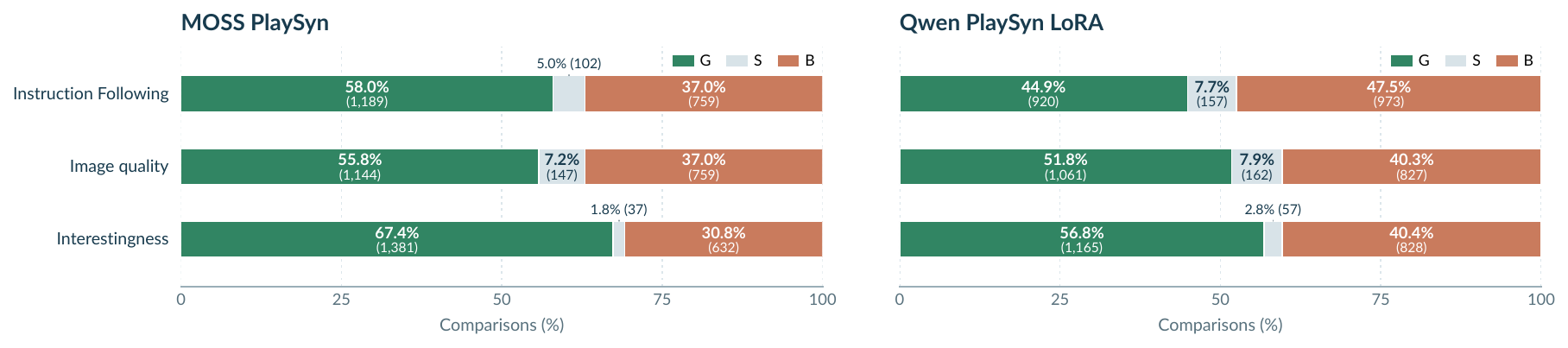}
  \caption{GSB comparison of our models with Qwen-Image-2.1 in instruction following, image quality, and interestingness. G and B denote preferences for our model and Base, respectively; S denotes a tie.}
  \label{fig:main_pairwise_gsb}
\end{figure*}

\section{PlayBench}
\label{sec:benchmark}

\subsection{Task Construction and Freezing}

\PlayBench contains 2,050 frozen tasks across 35 domains: 1,281 Core, 385 Challenge, and 384 Creative. Tasks contain one to four references: 1,747 single-reference, 279 two-reference, 22 three-reference, and two four-reference tasks. The benchmark covers 1,046 distinct identities, with held-out checks between training and evaluation.

Core measures common creative editing capabilities. Challenge emphasizes difficult conditions such as multiple references, identity, complex instructions, OCR, layout, and multiple people. Creative further emphasizes open-ended expression and interestingness. Every task follows the schema in \Cref{sec:creative_mining} and additionally fixes output dimensions and a random seed.

\subsection{Pointwise Evaluation}

The Pointwise judge receives the user instruction, references, and one model output. It evaluates every change and preservation constraint, and assesses identity, anatomy, OCR, copy-paste behavior, severe quality defects, and overall instruction score. For task $i$, let $a_i\in\{0,1\}$ indicate whether its output is accepted by the hard-quality gates:
\begin{equation}
  a_i=\mathbf{1}\left[\forall c\in\mathcal{C}_i,c=\mathrm{pass}\right]\cdot h_i,
\end{equation}
where $\mathbf{1}[\cdot]$ is the indicator function, $\mathcal{C}_i$ contains applicable change and preservation constraints, and $h_i=1$ if and only if all applicable hard checks for identity, anatomy, OCR, compositing, and quality pass; otherwise, $h_i=0$. The three-state label is Reject for any explicit failure, Uncertain when no explicit failure occurs but at least one item is uncertain, and Accept otherwise. Consequently, $a_i=1$ only for Accept.

The overall instruction score uses a 1--5 integer scale but cannot override a hard failure. Interestingness evaluation uses the dimensions in \Cref{sec:creative_eval}. As an auxiliary no-reference metric, we also report scores from aesthetic-predictor-v2-5~\cite{discus0434aesthetic}.

\subsection{Pairwise Blind Evaluation}

Pointwise evaluation tests absolute gates, whereas Pairwise evaluation asks which model is preferable on the same task. Each comparison receives identical references, instructions, and constraints, together with anonymous outputs from two models. Left--right positions are randomized per task, and the judge returns Left, Right, or Tie for instruction following, image quality, and interestingness.

For each comparison, we record Good, Same, and Bad (GSB): G means the PlaySyn model is preferred, S denotes a tie, and B means the Base model is preferred. We report the three counts and their percentages over all tasks. We also define the GSB score as
\begin{equation}
  \mathrm{GSB}=\frac{G+S}{B+S}.
\end{equation}
A score above 1 indicates more preferences for the PlaySyn model, while a score below 1 indicates more preferences for Base; higher is better.

\section{Experiments}
\label{sec:experiments}

\subsection{Models and Experimental Setup}

Our experiments include an unmodified Qwen-Image-2.1, \QwenPlay, and \MossPlay, providing a backbone baseline and testing PlaySyn training on two backbone families. We additionally evaluate MiniMax-H3-Image~\cite{minimax2026h3image} as an external editing baseline.

\begin{table*}[t]
  \centering
  \small
  \caption{Main PlayBench Pointwise results. Hard Pass is the hard-quality pass rate (\%); IF, Interestingness, and Aesthetic are mean scores.}
  \begin{tabular}{lcccc}
    \toprule
    Model & Hard Pass $\uparrow$ & IF $\uparrow$ & Interestingness $\uparrow$ & Aesthetic $\uparrow$ \\
    \midrule
    MiniMax-H3-Image & 73.82 & 4.4083 & \underline{3.5863} & 6.0173 \\
    Qwen-Image-2.1 & \underline{76.48} & \underline{4.4988} & 3.5395 & 5.9711 \\
    \midrule
    Qwen PlaySyn & 72.02 & 4.3787 & 3.5673 & \textbf{6.2177} \\
    MOSS PlaySyn & \textbf{78.51} & \textbf{4.5174} & \textbf{3.6251} & \underline{6.1968} \\
    \bottomrule
  \end{tabular}
  \label{tab:main_results}
\end{table*}

\paragraph{MOSS-Image-Edit-PlaySyn.} \Cref{fig:moss_playsyn_architecture} shows the model architecture. We initialize from MOSS-Image-Base~\cite{openmoss2026mossimage,mosi2026mossimage}. It has not undergone CFG distillation and retains conditional and unconditional prediction branches, allowing true CFG inference~\cite{ho2022classifierfree} and leaving more room for fine-tuning than a single-branch distilled model. MOSS-Image is natively text-to-image, so we add editing conditions following Qwen-Image-2.1~\cite{wu2025qwenimage,qwen2026image21}. Each reference follows two frozen encoding paths: the FLUX.2 VAE produces clean reference latents, while Qwen3.5-9B encodes the image and prompt into multimodal embeddings. After the DiT text projection, VAE reference tokens replace the corresponding VLM image-placeholder embeddings in place. The resulting context is concatenated with noisy target tokens and processed by the same 42 DiT blocks through joint self-attention. Target and text tokens use $\mathrm{Mod}(t)$, while clean reference tokens use $\mathrm{Mod}(0)$. Reference-specific 4D RoPE positions preserve input order and distinguish multiple images. Only the DiT is updated; the VAE and Qwen encoder remain frozen.

\paragraph{Two-stage training.} Stage 1 learns general editing capability. Starting from MOSS-Image-Base, we train on a mixture of an internally constructed version of ConceptEdit~\cite{cui2026conceptedit} and other internal editing data, covering object and attribute modification, style transfer, text rendering, and layout changes, with a learning rate of $2\times10^{-5}$. Stage 2 adapts to creative patterns: starting from the Stage 1 general-editing checkpoint, we perform full-parameter SFT on all 105,724 PlaySyn samples with a learning rate of $1\times10^{-5}$. The main experiment uses the 28,000-step checkpoint from this stage. MOSS uses a Flow Matching objective~\cite{lipman2023flowmatching} with logit-normal timestep sampling~\cite{esser2024rectifiedflow}; we use a shift of five and prompt dropout with probability 0.1 to maintain the unconditional branch required by true CFG inference.

\paragraph{Qwen-Image-2.1-PlaySyn.} Qwen-Image-2.1~\cite{qwen2026image21} is a CFG-distilled model and uses $\mathrm{CFG}=1$ by default. The current main results use Qwen PlaySyn LoRA~\cite{hu2021lora} trained on the \PlaySyn dataset, with rank 128, AdamW optimization, and a learning rate of $1\times10^{-4}$. It is loaded onto the original Qwen-Image-2.1 with a scale of 1.0 without modifying the Base weights.

All models generate on the same 2,050 tasks using identical user instructions, references, and per-task random seeds. The Qwen models and MOSS use 40 sampling steps. MOSS uses CFG 3.0; Qwen-Image-2.1 and Qwen PlaySyn use CFG 1.0. Both Qwen models retain the task-specified output dimensions, while MOSS follows the dynamic-resolution setting used in training and aligns dimensions to 16 pixels. MiniMax-H3-Image otherwise uses its own resolution and inference configuration. Cross-model results also reflect differences in initialization, training recipe, and inference configuration.
Automatic judging uses Qwen3.8-27B~\cite{qwen38}. Pointwise sample counts vary slightly because of a small number of service or parsing failures; the main Pairwise comparisons cover all 2,050 tasks.

\subsection{Pilot Ablation of CreativeBrief}
\label{sec:creative_brief_ablation}

We directly compare canonical-template synthesis with CreativeBrief synthesis, without training a downstream model. Both groups use the same tasks and references, and both go through PE fidelity checking and the same two-stage hard quality control. Judges view randomized left--right outputs and select the side that is more interesting and worth trying. There are 72 clean A/B pairs where both arms pass fidelity and hard quality requirements. CreativeBrief wins 32, the canonical baseline wins 10, and 25 are ties. In the other 5 pairs, judges find both outputs unacceptable in their interestingness judgments. Excluding ties and these five pairs gives a decisive preference win rate of $32/(32+10)=76.19\%$. \cref{sec:supp_ab} reports all 150 raw pairs and the clean subset.

\subsection{Main Results}

\Cref{tab:main_results} presents the main Pointwise results, and \Cref{fig:main_pairwise_gsb} shows the complete Pairwise GSB breakdown. MOSS PlaySyn achieves a hard-quality pass rate of 78.51\%, exceeding Qwen-Image-2.1 at 76.48\%, with higher mean instruction and interestingness scores. In complete Pairwise blind evaluation, MOSS PlaySyn attains GSB scores of 1.50, 1.42, and 2.12 over Base for instruction, quality, and interestingness, respectively.

Qwen PlaySyn LoRA behaves differently. Interestingness rises slightly ($3.5395\rightarrow3.5673$), but the hard-pass rate falls to 72.02\% and the instruction score to 4.3787. Pairwise instruction GSB is 0.95, while quality and interestingness GSB are 1.24 and 1.38. LoRA thus improves preference for quality and interestingness, but not absolute instruction or hard quality; the full-parameter MOSS model improves both. The auxiliary aesthetic scores are also higher for both PlaySyn-trained models: Qwen PlaySyn reaches 6.2177 and MOSS PlaySyn 6.1968, compared with 5.9711 for Qwen-Image-2.1.

The external baseline provides additional context. MiniMax-H3-Image reaches 73.82\% Hard Pass and 4.4083 IF, between Qwen-Image-2.1 and Qwen PlaySyn on these absolute metrics, while its Interestingness score of 3.5863 exceeds both Qwen variants. Its Aesthetic score is 6.0173; together with the other results, this shows that aesthetic quality alone does not determine constraint satisfaction or interestingness.

\begin{table}[t]
  \centering
  \small
  \caption{Identity, preservation, change, and OCR pass rates (\%).}
  \setlength{\tabcolsep}{2.5pt}
  \begin{tabular}{lcccc}
    \toprule
    Model & Identity & Preserve & Change & OCR \\
    \midrule
    MiniMax-H3-Image & 95.18 & 94.97 & \underline{97.75} & \underline{91.68} \\
    Qwen-Image-2.1 & \textbf{97.84} & \textbf{97.37} & 97.35 & 89.75 \\
    \midrule
    Qwen PlaySyn & 94.69 & 95.40 & 97.25 & 90.00 \\
    MOSS PlaySyn & \underline{95.43} & \underline{95.85} & \textbf{98.22} & \textbf{92.75} \\
    \bottomrule
  \end{tabular}
  \label{tab:consistency}
\end{table}

\begin{figure*}[t]
  \centering
  \includegraphics[width=\textwidth]{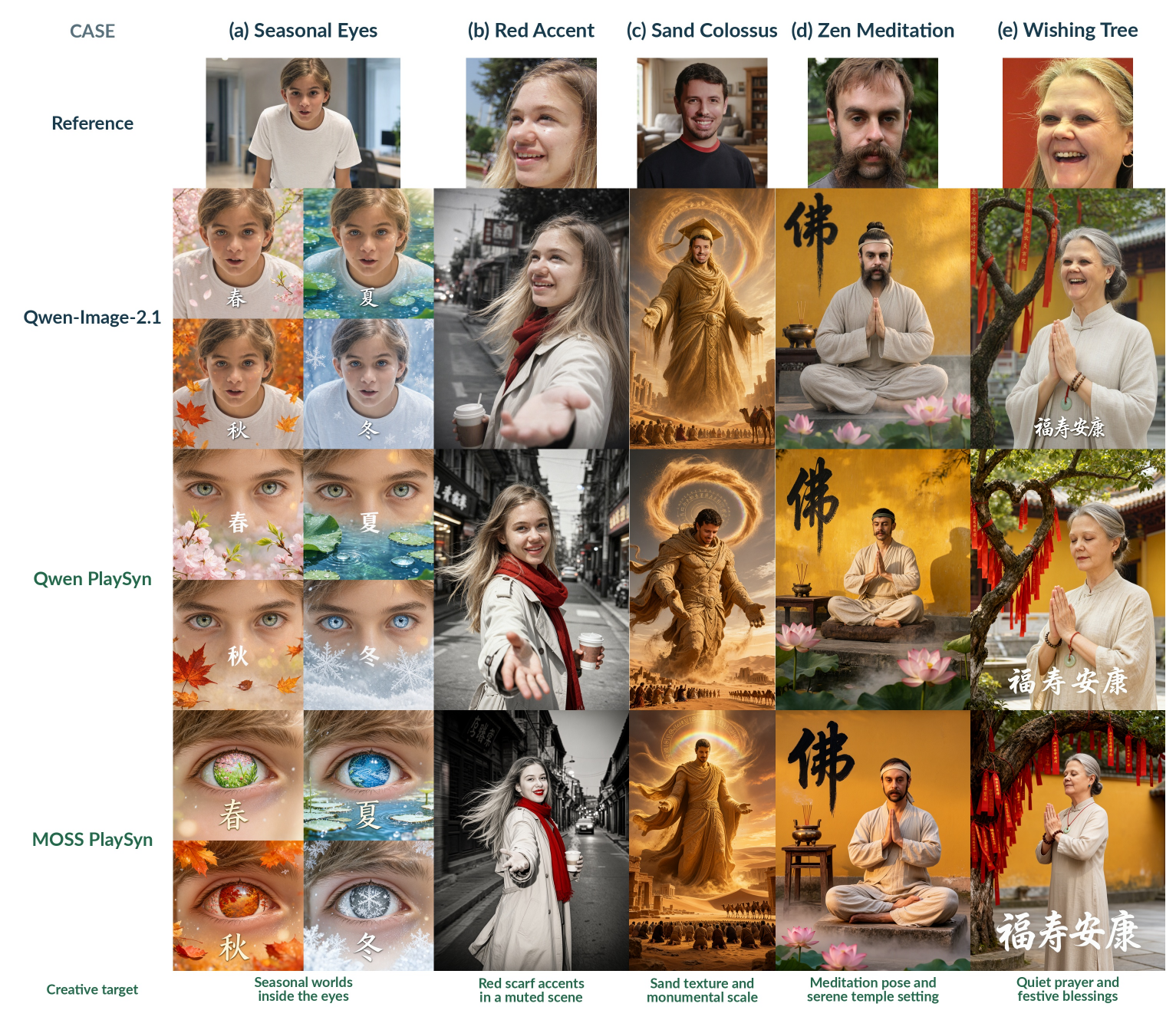}
  \caption{Visual comparison on PlayBench results. Rows show the input reference, Qwen-Image-2.1, Qwen PlaySyn, and MOSS PlaySyn. Bottom notes identify the creative mechanism highlighted by each task.}
  \label{fig:qwen_case_comparison}
\end{figure*}

\Cref{fig:qwen_case_comparison} compares both trained backbones on five Creative-track tasks. The examples emphasize different creative mechanisms: seasonal landscapes inside the eyes, a red scarf preserved within a muted scene, a figure transformed into a monumental sand deity, a meditation portrait, and a blessing scene. Each uses a single reference, while the two backbones produce distinct compositions, materials, color treatments, and narrative moods.

\subsection{Identity, Preservation, and Text Rendering}

MOSS PlaySyn completes more requested changes (98.22\% vs. 97.35\%) and renders text better (92.75\% vs. 89.75\%) in \Cref{tab:consistency}. It loses some identity and preservation scores (95.43/95.85 vs. 97.84/97.37). In the outputs we inspected, faces change more in pose, expression, and lighting; some of these changes are unintended. The model copies references less, but it also preserves them less reliably. MiniMax-H3-Image also performs well on Change and OCR (97.75/91.68).

\subsection{Complementarity of Pointwise and Pairwise Evaluation}

MOSS PlaySyn's Pointwise mean instruction score exceeds Base by 0.019, yet its Pairwise instruction GSB score reaches 1.50. Integer scores can hide differences in naturalness, detail, and constraint completion that become visible in a direct comparison. Similarly, an output can fail one preservation gate while still being preferable to an output with more severe errors. The two evaluations therefore answer different questions: Pointwise records constraint satisfaction, and Pairwise ranks outputs on the same task.

\section{Limitations}
\label{sec:limitations}

Our data and benchmark focus on portraits and real-world creative patterns, so the conclusions do not directly cover every general image-editing scenario. Interestingness is subjective, and VLM judgments can be affected by model bias. Local anatomical errors may also escape automatic checks, making human sampling and calibration with independent judges necessary.

\section{Conclusion}
\label{sec:conclusion}

We presented \PlaySyn, a data construction pipeline that mines recurring creative patterns from real user interactions and converts them into CreativeDNA and CreativeBriefs, producing 105,724 quality-controlled image-editing samples. We also introduced \PlayBench, a frozen benchmark of 2,050 tasks combining Pointwise constraint evaluation with randomized Pairwise comparison. We developed a multi-reference MOSS-Image editing architecture based on dual frozen encoding paths and in-place VAE-token replacement. Under the evaluated configurations, PlaySyn training improves VLM-judged Pairwise quality and interestingness on both model families, while MOSS PlaySyn additionally improves hard-quality pass rate and instruction following. The results show that reusable creative mechanisms mined from real interactions provide effective supervision and a principled basis for evaluating creative image editing.

{\small
\bibliographystyle{ieee_fullname}
\bibliography{PlaySyn}
}

\clearpage
\appendix
\makeatletter
\@addtoreset{table}{section}
\@addtoreset{figure}{section}
\@addtoreset{equation}{section}
\makeatother
\renewcommand{\thetable}{\Alph{section}.\arabic{table}}
\renewcommand{\thefigure}{\Alph{section}.\arabic{figure}}
\renewcommand{\theequation}{\Alph{section}.\arabic{equation}}
\renewcommand{\theHtable}{\Alph{section}.\arabic{table}}
\renewcommand{\theHfigure}{\Alph{section}.\arabic{figure}}
\renewcommand{\theHequation}{\Alph{section}.\arabic{equation}}
\begin{center}
  {\Large\bfseries Appendix\par}
\end{center}
\vspace{0.5em}

\section{Real-Data Use and Training Replacements}
\label{sec:supp_data_usage}

Real interactions are used only for pattern and creative mining. Training and evaluation use newly instantiated samples. \Cref{tab:data_usage_boundary} summarizes the role of each real-data source and its training replacement.

\begin{table}[!htbp]
  \centering
  \footnotesize
  \setlength{\tabcolsep}{2pt}
  \caption{Real-data use and training replacements.}
  \begin{tabular}{>{\raggedright\arraybackslash}p{0.24\columnwidth}>{\raggedright\arraybackslash}p{0.26\columnwidth}>{\raggedright\arraybackslash}p{0.15\columnwidth}>{\raggedright\arraybackslash}p{0.27\columnwidth}}
    \toprule
    Real data & Used for & In training? & Training replacement \\
    \midrule
    User images & Play mining & No & New references \\
    Interaction outputs & Creative mining & No & Generated targets \\
    User prompts & Intent mining & No & New instructions \\
    Statistics & Distribution analysis & No & -- \\
    \bottomrule
  \end{tabular}
  \label{tab:data_usage_boundary}
\end{table}

\section{Technical Details of PlaySpec and CreativeDNA}
\label{sec:supp_schema}

\paragraph{Real-prototype candidate pool.} Retrieval has two steps: build a shared pool, then sample prototypes for each task. Each pattern contributes at most 10 real samples. Samples are grouped by their subcategory before fine clustering and selected round-robin, so no subcategory can fill the quota by itself. The pool contains 565,878 samples from 104,428 creative patterns and 113,539 such subcategories. Each synthesis task draws up to three prototypes from this ordered pool for CreativeDNA extraction. A prototype stores the real input references and output image, raw prompt, detailed edit description, pattern name before fine clustering, and cluster ID. Real images are used only to analyze creative mechanisms. They are not used as references for synthetic tasks or included in released data. CreativeBrief reads structured DNA and text descriptions of new references; it does not read real user images.

\paragraph{CreativeDNA fields and validation.} \Cref{tab:supp_dna_schema} lists the complete schema. Records are rejected if fields are missing, arrays are empty, types are invalid, or scores fall outside their allowed ranges. Quality words such as high-definition, premium, beautiful, or atmospheric are not visual hooks by themselves. Creative records must identify a visible point of interest and an event in the image. Utility records must state their practical value. Face swapping and identity replacement are allowed, but merging several human identities into one person is not.

\begin{table}[t]
  \centering
  \small
  \caption{Complete CreativeDNA fields and their semantics.}
  \setlength{\tabcolsep}{4pt}
  \begin{tabular}{p{0.31\columnwidth}p{0.61\columnwidth}}
    \toprule
    Field & Meaning and requirement \\
    \midrule
    \field{play\_type} & \field{creative} or \field{utility}, distinguishing creative from practical tasks \\
    \field{visual\_hook} & The immediately visible core point of interest; concrete rather than a quality adjective \\
    \field{contrast} & Contrast in scale, material, era, reality--fantasy, or related dimensions \\
    \field{story} & The event taking place in the image \\
    \field{subject\_}\allowbreak\field{relation} & Relationships among people, objects, and characters \\
    \field{signature\_}\allowbreak\field{elements} & Array of signature visual elements forming the mechanism \\
    \field{composition} & Distinctive layout, viewpoint, or framing \\
    \field{emotional\_}\allowbreak\field{payoff} & Emotional value such as surprise, humor, comfort, or awe \\
    \field{required\_}\allowbreak\field{operations} & Editing operations required to realize the mechanism \\
    \field{reference\_}\allowbreak\field{requirements} & Content and role requirements for input references \\
    \field{utility\_note} & Practical-value explanation for Utility tasks \\
    \field{interest\_score} & Interestingness on a 1--5 scale \\
    \bottomrule
  \end{tabular}
  \label{tab:supp_dna_schema}
\end{table}

\paragraph{CreativeBrief fields and generation modes.} Each Brief contains \field{creation\_mode}, \field{visual\_hook}, \field{story\_scene}, \field{signature\_elements}, \field{composition}, \field{emotional\_payoff}, \field{creative\_prompt}, and \field{additional\_constraints}. The last field stores \field{description} and \field{refs} for new visible requirements beyond the base constraints. Each requirement must be atomic and checkable. A Creative task produces two Briefs. Exemplar reuses the core mechanism of one prototype; Within-play Recombine combines compatible mechanisms from several prototypes of the same play. A Utility task produces two practical proposals without adding novelty that would break the task. The two Briefs must differ in mechanism, story, or composition, not just wording.

\section{CreativeDNA and CreativeBrief Conversion Examples}
\label{sec:supp_examples}

\paragraph{Cherry-blossom sports car: from images to CreativeDNA.} In this real sample, a person drives a futuristic convertible covered in cherry blossoms along a flower-lined road at sunset. The mechanism is:
\begin{quote}
\small
\textbf{visual\_hook:} A hard metallic sports car is thickly covered in soft pink blossoms.\\
\textbf{contrast:} Futuristic machinery contrasts with delicate spring flowers.\\
\textbf{story:} The character drives through a scene that combines science fiction and romance.\\
\textbf{composition:} A low three-quarter view uses the road as a leading line and the sunset as a distant focal point.\\
\textbf{signature\_elements:} A futuristic convertible, blossoms covering its body, an arcade of cherry trees, and a warm sunset.
\end{quote}
We reuse the contrast between hard and soft elements, the driving action, and the leading-line composition. We do not copy the entire cherry-blossom car into a new task.

\paragraph{Giant rose portrait: from CreativeBrief to a verifiable instruction.} The original task preserves a person, creates a giant rose portrait of that person, and shows both together. Simply expanding the operation template could produce an ordinary person standing in front of a painting. A stronger creative version is:
\begin{quote}
\small
\textbf{visual\_hook:} The real person looks up at a giant self-portrait on a stone wall; red roses bloom from the depicted person's shoulders, and vines hang from the frame onto the physical ground.\\
\textbf{story\_scene:} In a castle at dusk, the person appears to attend a rose celebration held in their own honor.\\
\textbf{composition:} A low-angle ultra-wide view with the real person in the lower left and the giant portrait on the stone wall to the right, creating a scale contrast.\\
\textbf{emotional\_payoff:} Romance, fairy-tale wonder, and the surprise of being crowned by roses.
\end{quote}
The corresponding user instruction is:
\begin{quote}
\small
Extract the person from \texttt{<Picture 1>} and preserve the facial features, hairstyle, and clothing. Place the person in a rose-castle courtyard at dusk. Create a giant rose portrait of the person on the stone wall to the right, replacing the depicted shoulder and neck area with blooming red roses. Let vines extend from the lower edge of the frame onto the stone-paved ground. The real person in the foreground looks slightly upward and appears together with the complete giant portrait. Use a low-angle ultra-wide composition and a realistic cinematic style.
\end{quote}
Three checkable constraints are added: roses surround the shoulders and neck without covering the face; vines extend visibly from the frame to the stone ground; and both the real person and the complete giant portrait remain visible. These constraints let quality control test the scale contrast, rose substitution, frame-breaking effect, and encounter between the person and their depicted self.

\section{PE Fidelity, Candidate Generation, and Quality Gates}
\label{sec:supp_pe}

\paragraph{PE fidelity checking.} The judge reads the pattern name, original instruction, PE rewrite, base constraints, preserved attributes, reference roles, new creative constraints, and all reference images together. Original edits, preserved attributes, identity sources, specified text, key counts, and spatial relationships are hard constraints. The rewrite may add lighting, material, or minor details only when they agree with the original intent. Keeping text, clothing, accessories, people, or background elements already visible in a reference is allowed. Rejection labels include \field{dropped\_constraint}, \field{changed\_identity\_role}, \field{changed\_text}, \field{contradiction}, \field{unsupported\_addition}, and \field{identity\_blend}. A rewrite is not rejected simply for being longer or more detailed. If the images do not provide enough evidence, the result is marked uncertain.

\paragraph{Minimal repair and version checking.} Failed rewrites can be repaired or regenerated with PE. Repair changes only the wording needed to restore the original instruction, hard constraints, identity sources, specified text, and spatial relationships. It adds no new creative pattern and leaves the original training prompt unchanged. Every repaired version must pass the image-based fidelity check again. Before generation, the pipeline compares the instruction hash with the inspected version. Generation is blocked if the check is missing, outdated, rejected, or uncertain.

\paragraph{Candidate and filtering order.} Each task produces two Briefs and uses two different seeds per Brief, giving four candidates. We check changes and preserved attributes, identity and copy-paste behavior, visual quality and anatomy, and OCR. A second independent review looks for remaining failures. Interestingness is scored only after all applicable hard checks pass. It ranks passing candidates and cannot compensate for identity or core-constraint failures.

\clearpage
\onecolumn
\section{Canonical Templates versus CreativeBrief}
\label{sec:supp_cases}

\Cref{tab:supp_template_creative_cases} compares six representative task pairs. CreativeBrief adds specific visual mechanisms, spatial relationships, or events, then turns them into checkable constraints. It does more than add decorative adjectives.

\begin{table}[!htbp]
  \centering
  \footnotesize
  \caption{Representative canonical-to-CreativeBrief differences. \CBhook{Teal} marks creative mechanisms; \CBdetail{blue} marks checkable spatial or structural details.}
  \setlength{\tabcolsep}{5pt}
  \begin{tabular}{p{0.12\textwidth}p{0.34\textwidth}p{0.46\textwidth}}
    \toprule
    Play pattern & Canonical baseline & New CreativeBrief mechanisms and checkable details \\
    \midrule
    Christmas reclining portrait & Christmas dress and hat, Christmas tree, gift boxes, and a few snowflake stickers & Snowman, gingerbread-person, and red-heart 2D stickers; a \CBhook{white hand-drawn dashed outline} around the head; \CBdetail{tree and gifts placed separately on the left and right} \\
    Resin figurine & 1/7-scale figurine, circular base, desk, coffee, old books, and headphones & A \CBdetail{complete full-body figurine that can stand}; a \CBhook{monitor showing both gray-model and color renders}; \CBhook{packaging printed with the same character} \\
    Two-person fist bump & Standing side by side, central fist bump, paired rings, and FRIENDS / FOREVER text & \CBdetail{Fist-bump point in the lower third}; \CBhook{two rings forming a continuous arc of light}; blurred upper bodies as a soft background \\
    Three-person studio group & Overhead view, three people side by side, specified clothing and expression edits & \CBdetail{Back-left, front-center, and back-right layering} under a steep overhead view; \CBhook{grid shadows with consistent directions on the floor}; \CBdetail{all three looking slightly upward at the camera} \\
    Real person and cartoon double & Matching flat cartoon double on a graffiti wall with Happy Graffiti text & \CBhook{The double waves, winks, and appears to emerge from the wall}; \CBdetail{spray cans and colored dots at the wall base} create a \CBhook{street-photo event} \\
    Snow self-portrait & A line-drawn self-portrait made in snow with a wooden shovel, New Year text, and a snowman & \CBdetail{Black line drawing in the front left matching the person's facial features}; \CBdetail{handwritten text to the right of the portrait}; a \CBdetail{red-hatted snowman at the lower right}; \CBhook{wooden door and warm string lights} \\
    \bottomrule
  \end{tabular}
  \label{tab:supp_template_creative_cases}
\end{table}

\section{Human CreativeBrief A/B Pilot}
\label{sec:supp_ab}

The pilot uses 200 existing tasks, 134 creative patterns, and 594 real creative prototypes to compare CreativeBrief with the canonical pipeline. Both arms share the same tasks and references. Judges choose the side that is more interesting and worth trying; they can also select a tie or mark both outputs as unacceptable in terms of interestingness. The raw set has 150 pairs. The clean set keeps 72 pairs in which both sides receive keep in the latest PE check and pass the same two-stage hard quality control. The decisive win rate is $W_{\mathrm{creative}}/(W_{\mathrm{creative}}+W_{\mathrm{baseline}})$, excluding ties and mutually unacceptable pairs.

\begin{table}[!htbp]
  \centering
  \footnotesize
  \caption{Human blind A/B evaluation of CreativeBrief and the canonical baseline.}
  \setlength{\tabcolsep}{3.5pt}
  \begin{tabular}{@{}lrrrrrr@{}}
    \toprule
    Evaluation set & Pairs & Creative wins & Baseline wins & Ties & Both unacceptable & Decisive win rate \\
    \midrule
    Raw A/B & 150 & 70 & 19 & 48 & 13 & 78.65\% \\
    Clean A/B (PE + hard QC) & 72 & 32 & 10 & 25 & 5 & 76.19\% \\
    \bottomrule
  \end{tabular}
  \label{tab:supp_creative_ab}
\end{table}

We report the raw and clean sets together. In the clean set, CreativeBrief wins more often, suggesting that its appeal is not solely an artifact of PE drift or differences in hard-gate eligibility.

\end{document}